\documentclass[conference]{IEEEtran}

\IEEEoverridecommandlockouts                              

\usepackage{enumitem}
\setlist{nolistsep}
\usepackage{multirow}
\usepackage{array}
\usepackage{amsmath}
\usepackage{amssymb}
\usepackage{amsthm}
\usepackage{graphicx}
\usepackage{color}
\usepackage{lscape}
\usepackage{float}
\usepackage[table]{xcolor}
\usepackage{nidanfloat}
\usepackage{cite}

\newfloat{matrixeqn}{t!}{matrixequation}

\newtheorem{definition}{Definition}

\newcommand{\R}{\mathbb{R}}

\newcommand{\bfa}{{\bf a}}

\newcommand{\bfc}{{\bf c}}

\newcommand{\bfe}{{\bf e}}

\newcommand{\cA}{ {\cal A}}
\newcommand{\cB}{ {\cal B}}
\newcommand{\cP}{ {\cal P}}
\newcommand{\cU}{ {\cal U}}
\newcommand{\cX}{ {\cal X}}
\newcommand{\cC}{ {\cal C}}

\newcommand{\cD}{ {\cal D}}
\newcommand{\cI}{ {\cal I}}
\newcommand{\cK}{ {\cal K}}

\newcommand{\bea}{ \left[ \begin{array} }
\newcommand{\eea}{ \end{array} \right] }
\newcommand{\circu}{ \mbox{\tt bcirc} }

\newcommand{\fld}{ \mbox{\tt fold}}

\newcommand{\matvec}{ \mbox{\tt MatVec}}

\newcommand {\by}	    {\times}
\newcommand {\mat}	    {\begin{bmatrix}}
\newcommand {\rix}	    {\end{bmatrix}}

\newcommand{\mathR}{\mathbb{R}^{\ell \times m \times n}}

\author{Randy C. Hoover$^{1}$, Kyle Caudle$^{2}$ and Karen Braman$^{2}$
\thanks{*The current research was supported in part by the Department of the Navy, Naval Engineering Education Consortium under Grant No. (N00174-19-1-0014), the NASA Space Grant Consortium and the National Science Foundation under Grant No. (2007367).  Any opinions, findings, and conclusions or recommendations expressed in this material are those of the authors and do not necessarily reflect the views of the Naval Engineering Education Consortium, NASA or the National Science Foundation.}
\thanks{$^{1}$Randy C. Hoover is with the department of Computer Science and Engineering,
        South Dakota Mines, Rapid City, SD, USA
        {\tt\small randy.hoover@sdsmt.edu}}%
\thanks{$^{2}$Kyle Caudle and Karen Braman the Department Mathematics, South Dakota Mines, Rapid City, SD, USA
        {\tt\small \{kyle.caudle,karen.braman\}@sdsmt,edu}}%
}

\begin{document}
%
\title{A New Approach to Multilinear Dynamical Systems and Control*}



%


\maketitle
\thispagestyle{empty}
\pagestyle{empty}


\begin{abstract}
The current paper presents a new approach to multilinear dynamical systems analysis and control.  The approach is based upon recent developments in tensor decompositions and a newly defined algebra of circulants.  In particular, it is shown that under the right tensor multiplication operator, a third order tensor can be written as a product of third order tensors that is analogous to a traditional matrix eigenvalue decomposition where the ``eigenvectors'' become eigenmatrices and the ``eigenvalues'' become eigen-tuples.  This new development allows for a proper tensor eigenvalue decomposition to be defined and has natural extension to linear systems theory through a \textit{tensor-exponential}.  Through this framework we extend many of traditional techniques used in linear system theory to their multilinear counterpart.
\end{abstract}

\section{Introduction} \label{sec:intro}
Traditional approaches to the analysis and control of linear time invariant (LTI) systems is well known and well understood.  However as systems become increasingly complex, and multi-dimensional measurement devices become more commonplace, extensions from the linear system to a multilinear system framework needs to be developed.  While there have been several approaches developed to investigate multilinear dynamical systems, most rely on decompositions revolving around either the Tucker or Canonical Decomposition/Parallel Factors (commonly referred to collectively as the CP decomposition).  Tucker/CP provides a framework for decomposing a high order \textit{tensor} into a collection of factor matrices multiplying a ``core tensor".  The structure of the core-tensor depends on which factorization strategy is being used (Tucker produces a ``dense" core whereas CP produces a ``diagonal" core).  Regardless of the decomposition being applied, both are regarded as form of higher order singular value decomposition~\cite{art:Tucker66, art:Harshman70, art:Lathauwer00,art:Kolda09}.

As a form of high-order singular value decomposition, Tucker/CP algorithms have a natural fit within the machine learning community where data naturally arises as two-dimensional structures, e.g. digital image data.  As such, these algorithms have played a central role in extending many of the existing machine learning algorithms~\cite{art:Tucker66, art:Harshman70, art:Lathauwer00,art:Alex02, art:Vasilescu07_ICCV,art:Vasilescu03,art:MSLSurvey2011,art:Hoover10,art:Hoover11,art:Hao13,art:Kilmer13,art:Kolda09}.  However, as their popularity has gained more traction over the last decade, they have made their way into the dynamical systems and controls community as well~\cite{art:Sun06,art:Weijun18,Kruppa2018FeedbackLO,KRUPPA20149474,MULLER2015416,DBLP:conf/simultech/PangalosEL13a,art:Sewe17,DBLP:conf/simultech/PangalosEL13,DBLP:conf/IEEEcca/PfeifferLSS12,BAIER20175630,552290,NIPS2013_5117,Gel2018MultidimensionalAO}.  While most applications of Tucker/CP in the dynamical systems and controls community revolve around the reduction of certain classes of nonlinear systems to multilinear counterparts~\cite{Kruppa2018FeedbackLO,KRUPPA20149474,MULLER2015416,DBLP:conf/simultech/PangalosEL13a,art:Sewe17,DBLP:conf/simultech/PangalosEL13,DBLP:conf/IEEEcca/PfeifferLSS12,Gel2018MultidimensionalAO}, others have focused on time-series modeling~\cite{Lu2018TransformBasedMD,NIPS2013_5117}, fuzzy inference~\cite{552290}, or identification/modeling of inverse dynamics~\cite{BAIER20175630}.

In the current paper, we describe a new approach to multilinear dynamical systems analysis and control through Fourier theory and an algebra of circulants as outlined in~\cite{KilMP08,KilM09,Bra10,KilBH11,art:David11}.  It is shown that under the right tensor multiplication operator, a third order tensor can be written as a product of third order tensors in which the left tensor is a collection of eigenmatrices, the middle tensor is a front-face diagonal (denoted as f-diagonal) tensor of eigen-tuples, and the right tensor is the tensor inverse of the eigenmatrices resulting in a tensor-tensor eignevalue decomposition that is similar to its matrix counterpart.  Moreover, using the aformentioned decomposition,~\cite{art:Lund18,th:Lund18} illustrates that a multilinear system of ordinary differential equations (MODEs) can be effectively solved via a tensor-version of the matrix exponential (referred to as the t-exponential).  

Building on the work of~\cite{KilMP08,KilM09,Bra10,KilBH11,art:David11,art:Lund18,th:Lund18}, the contributions of the current paper are four fold: (1) we extend the results of~\cite{art:Lund18,th:Lund18} to include the zero-state response to the multilinear dynamical system in an effort to introduce multilinear feedback control, (2) we develop a stability criterion for the multilinear dynamical system to include exponential convergence of system trajectories, (3) we introduce a new approach to validate controllability of multilinear systems using a block-Krylov subspace condition, and finally (4) we present a method to design multilinear state-feedback control using the developments of (1) - (3). 

The remainder of this paper is organized as follows:  In Section~\ref{sec:math_prelims} we discuss the relevant tensor algebra and the newly defined tensor multiplication operator.  In Section~\ref{sec:t_func} we present the tensor-tensor eigenvalue decomposition and show how it can be used to define functions on tensors (namely the tensor exponential).  In Section~\ref{sec:MLTIsys} we provide several extensions of traditional linear systems theory to their multilinear counterpart. Section~\ref{sec:example} we provide an illustrative example of the newly developed theory and finally, Section~\ref{sec:conclusions} presents some discussion and provides some insight into future research directions.

\section{Mathematical Foundations of Tensors} \label{sec:math_prelims}
In the current section we discuss the mathematical foundations of the tensor decompositions used in the current work.  While most of the theory in this section is outlined in~\cite{KilMP08,KilM09,Bra10,art:Hoover11,art:Lund18,th:Lund18}, we summarize this theory here to keep the current work self contained.

The term \emph{tensor}, as used in the context of this paper, refers to a
multi-dimensional array of numbers, sometimes called an \emph{n-way} or
\emph{n-mode} array.  If, for example, $\cA \in \R^{\ell \by m \by n}$ then
we say $\cA$ is a third-order tensor where \emph{order} is the number of ways
or modes of the tensor.  Thus, matrices and vectors are second-order and
first-order tensors, respectively.  
Fundamental to the results presented in this paper is a recently defined
multiplication operation on third-order tensors which itself produces a
third-order tensor~\cite{KilMP08,KilM09}.  

Further, it has been shown in~\cite{Bra10} that under this multiplication operation, $\mathR$ 
is a free module over a commutative ring with unity where the ``scalars" are $\R^{1 \times 1 \times n}$ tuples. In addition, 
it has been shown in~\cite{Bra10} and~\cite{KilBH11} that all linear
transformations on the space $\mathR$ can be represented by
multiplication by a third-order tensor.  Thus, even though $\mathR$ is not strictly a vector space, many of the
familiar tools of matrix linear algebra can be applied in this
new context, including the basic building blocks for dynamical systems and control of multilinear systems.  For a more in depth discussion on this topic, the reader is referred to~\cite{Bra10}.

First, we review the basic definitions from
\cite{KilM09} and \cite{KilMP08} and introduce some basic notation.
It will be convenient to break a tensor $\cA$ in $\mathR$ up into
various slices and tubal elements, and to have an indexing on those.
The $i^\text{th}$ lateral slice will be denoted $\cA_i$ whereas the $j^\text{th}$
frontal slice will be denoted $\cA^{(j)}$.  In terms of {\sc Matlab}
indexing notation, this means $\cA_i \equiv \cA(:,i,:)$ 
while $\cA^{(j)} \equiv \cA(:,:,j)$.

We use the notation $\bfa_{ik}$ to denote the $i,k^\text{th}$ tube in$\cA$;
that is $\bfa_{ik} = \cA(i,k,:)$.  The $j^\text{th}$ entry in that tube is
$\bfa_{ik}^{(j)}$.  Indeed, these tubes have special meaning for us in 
the present work, as they will play a role similar to scalars in $\mathbb{R}$.
Thus, we make the following definition:

\begin{definition} An element $\bfc \in \mathbb{R}^{1 \times 1 \times n}$ is
called a {\bf tubal-scalar} of length $n$.  
\end{definition}

As mentioned previously, the set of tubal-scalars with length $n$ endowed with element-wise addition and
tensor multiplication (defined by the \textit{t-product} in Def. 2) forms a commutative ring~\cite{Bra10}.  For ease of notation, we will use ${\bf 0}$ to denote the additive identity, i.e., the
tubal-scalar with all zero elements. Let $\bfe_1$ denote the tubal-scalar with
all zero elements except a $1$ in the first position. Then it is easy to see
that $\bfe_1$ is the multiplicative identity in this ring and it will play an
important role in the remaining tensor definitions.

In order to discuss multiplication between two tensors we must first introduce
the concept of converting $\cA \in \mathR$ into a block circulant matrix.

If $\cA \in \mathR$ with $\ell \times m$ frontal slices then 
\[ \circu ( \cA) = \bea{ccccc}  A^{(1)} & A^{(n)} & A^{(n-1)} & \ldots & A^{(2)} \\ 
                                A^{(2)} & A^{(1)} & A^{(n)} & \ldots & A^{(3)} \\
                                \vdots & \ddots & \ddots & \ddots & \vdots \\
                             A^{(n)} & A^{(n-1)} & \ddots & A^{(2)} & A^{(1)} \eea, \]
is a block circulant matrix of size $\ell n \times m n$.   

We anchor the $\matvec$ command to the frontal slices of the tensor.
$\matvec(\cA)$ takes an $\ell \times m \times n$ tensor and returns a
block $\ell n \times m$ matrix \[ \matvec(\cA) = \bea{c} A^{(1)} \\
A^{(2)} \\ \vdots \\ A^{(n)} \eea.\] The operation that takes
$\matvec(\cA)$ back to tensor form is the $\fld$ command: \[ \fld(
\matvec (\cA) ) = \cA .\]

With these two operations in hand, we introduce the \textit{t-product} between two, third-order tensors~\cite{KilMP08,KilM09}:
\vspace{0.1cm}
\begin{definition} \label{def:mult}
Let $\cA \in \R^{\ell \times p \times n}$ and $\cB \in \R^{p \times m \times n}$ be two third order tensors.  Then
the \textbf{t-product} $\cA * \cB \in \R^{\ell \times m \times n}$ is defined as 
\[ \cA * \cB = \fld \left( \circu( \cA ) \cdot\matvec (\cB)~\right).\]
\end{definition}~\\
Note that the tensor \textit{t-product} enables the multiplication of two third order tensors via \textit{mod-n} circular convolution.  Moreover, in general, the \textit{t-product} of two tensors will not commute, with the exception in which
$\ell=p = m = 1$, i.e., when the tensors are tubal-scalars.  As a matter of illustration, \textit{Example 1} details the application of the \textit{t-product} on two third order tensors.\\~\\
\textit{Example 1:}  Suppose $\cA\in\R^{\ell \times p \times 3}$ and $\cB\in\R^{p \times m \times 3}$.
Then
\[
\cA*\cB=\fld \left(\left[\begin{array}{ccc}
                A^{(1)} & A^{(3)} & A^{(2)} \\ A^{(2)} & A^{(1)} & A^{(3)} \\ A^{(3)} 
			& A^{(2)} & A^{(1)} \end{array}\right]
         \left[\begin{array}{c}
                B^{(1)} \\ B^{(2)}\\ B^{(3)} \end{array}\right] \right).
\]

\begin{definition}\label{def:i} The \textbf{identity tensor}
$\cI \in \R^{m\by m\by n}$ is the tensor whose frontal slice is the $m \times m$ identity
matrix, and whose other frontal slices are all zeros.  
\end{definition}

\begin{definition} \label{def:trans}
If $\cA$ is $\ell \times m \times n$, then the \textbf{tensor transpose} $\cA^T$ is 
the $m \times \ell \times
n$ tensor obtained by transposing each of the frontal slices and then reversing
the order of transposed frontal slices 2 through $n$.  
\end{definition}

\begin{definition}\label{def:inverse} 
A tensor $\cA \in \R^{n\by n\by \ell}$ has an \textbf{tensor inverse} $\cB  \in \R^{n\by n\by \ell}$ provided
$$
	\cA * \cB = \cI \text{  and  } \cB * \cA = \cI
$$
where $\cI \in \R^{n \times n \times \ell}$.  The tensor inverse is computed as  
$$
	\cA^{-1} = \fld(\circu(\cA)^{-1}).
$$
\end{definition}

\section{Tensor Eigenvalue Decomposition and Functions of Tensors}\label{sec:t_func}
In this section we present the tools required to extend traditional linear systems theory and control to their multilinear systems domain.  Namely, the computation of an \textit{eigenvalue-like} decomposition for third order tensors.  Such a decomposition provides a natural interpretation to defining functions of tensors~\cite{th:Lund18,art:Lund18}, canonical forms~\cite{art:Miao19}, and multilinear time-series analysis~\cite{art:Weijun18}.

\subsection{Computation of the t-eigenvalue decomposition}
\begin{figure*}[b]
\begin{center}
\includegraphics[width=0.9\textwidth]{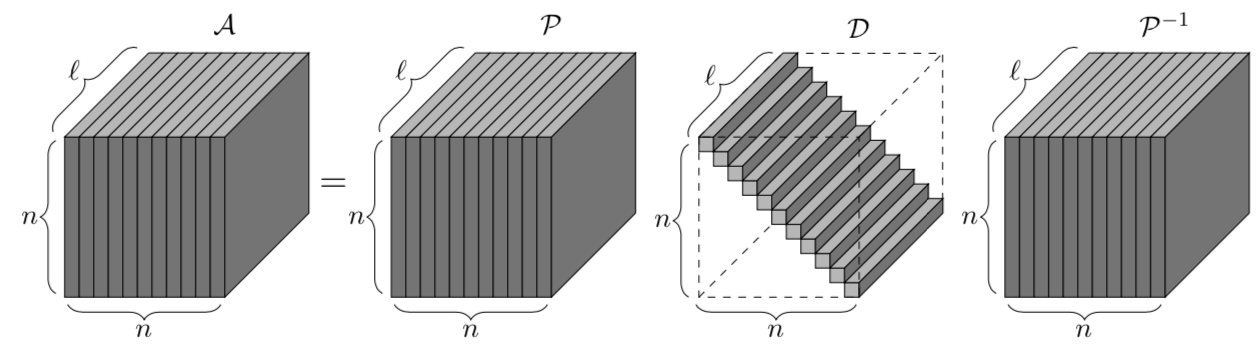}
\end{center}
\caption{\label{fig:svd}Graphical illustration of the \textbf{t-eig} of an $n \by n \by \ell$ tensor.  
}
\end{figure*}

In an effort to extend traditional linear-time invariant systems analysis and control to their multilinear counterparts, we present the tensor-tensor eigenvalue decomposition.  In~\cite{Bra10,art:Hao13,art:Kilmer13} the authors show that, for $\cA \in \mathbb{R}^{n \times n \times \ell}$, there exists an $n \times n \times \ell$ tensor $\mathcal{P}$ and an $n \times n \times \ell$ f-diagonal (front-face diagonal) tensor $\mathcal{D}$ such that 
\begin{equation} 
\cA = \mathcal{P} * \mathcal{D} * \mathcal{P}^{-1} \implies \cA * \mathcal{P} = \mathcal{D} * \mathcal{P} \implies \cA * \mathcal{P}_j = \mathcal{P}_j \textbf{d}_j.
\label{eq:t-eig}
\end{equation}
Moreover, the overall ``structure" of the deomposition is similar to a matrix eigenvalue decomposition in that we get a new tensor $\cP$ who's lateral slices are analogous to eigenvectors (referred to as eigenmatrices) and an f-diagonal tensor $\cD$ who's ``tubal scalars" $\textbf{d}_j = \mathcal{D}(j,j,:)$ are analogous to eigenvalues (referred to as eigentuples).  Throughout this paper, we refer to this decomposition as the \textbf{t-eig}, a graphical illustration of the which is shown in Figure~\ref{fig:svd}.

Computation of the \textbf{t-eig} comes from the constructive proof outlined in~\cite{Bra10,art:Hao13,art:Kilmer13}, that will be restated here for completeness.  It is well known in matrix theory that a circulant matrix can be diagonalized via left and right multiplication by a discrete Fourier transform (DFT) matrix.  Similarly, a block circulant matrix can be block diagonalized via left and right multiplication by a block diagonal DFT matrix.  For example, consider the tensor $\cA \in \mathbb{R}^{n \times n \times \ell}$, then
\begin{equation}
	(F_n \otimes I_\ell) \circu(\cA) (F_n^* \otimes I_\ell)
	 = \left[
	 \begin{array}{cccc}
      	D_1 &  &  &  \\
      	& D_2 & & \\
      	&& \ddots &\\
      	&&& D_\ell
     \end{array}
     \right],
     \label{eq:D}
\end{equation}
where each of the $D_i$ are $n \times n$, $I$ is an $\ell \times \ell$ identity matrix, $F_n$ is the $n \times n$ DFT matrix, 
\begin{equation}
     F_n = \frac{1}{\sqrt{n}} \left[
	 \begin{array}{ccccc}
	    1 & 1 & 1 & 1 & 1\\
      	1 & \omega & \omega^2 & \cdots & \omega^{n-1}\\
      	1 & \omega^2 & \omega^4 & \cdots & \omega^{2(n-1)}\\
      	\vdots & \vdots & \vdots & \ddots & \vdots\\
      	1 & \omega^{n-1} & \omega^{2(n-1)} & \cdots & \omega^{(n-1)(n-1)}\\
     \end{array}
     \right],
	\label{eq:DFT}
\end{equation}
where $\omega = e^{-2 \pi i/n}$ is a primitive $n^\text{th}$ root of unity, $F_n^*$ is its conjugate transpose, and $\otimes$ is the Kronecker product.  To construct the \textbf{t-eig} defined in~(\ref{eq:t-eig}), the matrix eigenvalue decomposition is performed on each of the $D_i$, i.e., $D_i = P_i \Lambda_i P_i^{-1}$ resulting in the decomposition\\   
$
	\left[
	 \begin{array}{ccc}
      	D_1 &  & \\
      	& \ddots &\\
      	&& D_\ell
     \end{array}
     \right] =
$
\begin{equation}
     \left[
	 \begin{array}{ccc}
      	P_1 &  & \\
      	& \ddots &\\
      	&& P_\ell
     \end{array}
     \right]
     \left[
	 \begin{array}{ccc}
      	\Lambda_1 &  & \\
      	& \ddots &\\
      	&& \Lambda_\ell
     \end{array}
     \right]
     \left[
	 \begin{array}{ccc}
      	P_1^{-1} &  & \\
      	& \ddots &\\
      	&& P_\ell^{-1}
     \end{array}
     \right].
	\label{eq:diag}
\end{equation}

Applying $(F_n^* \otimes I_n)$ to the left and $(F_n \otimes I_n)$ to the right of each of the block diagonal matrices on the right hand side of~(\ref{eq:diag}) results in each being block circulant, i.e., if we define $\hat{P}$ as the block diagonal matrix with $P_i$ as its diagonal blocks, then
\[
	(F_n^* \otimes I_\ell) \hat{P} (F_n \otimes I_\ell) = \left[
     \begin{array}{cccc}
      	P_1 & P_\ell & \cdots & P_{\ell-1}\\
      	P_2 & P_1& \cdots & P_{\ell-2}\\
      	\vdots & \vdots & \ddots & \vdots\\
      	P_\ell & P_{\ell-1} & \dots & P_1
     \end{array}
     \right].
\]
Taking the first block column of each block circulant matrix and applying the $\texttt{fold}$ operator results in the decomposition $\mathcal{P} * \mathcal{D} * \mathcal{P}^{-1}$.  Note that for simplicity, as well as computational efficiency, this entire process can by performed using the fast Fourier transform in place of the DFT matrix as illustrated in~\cite{KilMP08, KilM09,Bra10,art:Hao13,art:Kilmer13}.

\subsection{Functions of tensors}
The results of the preceding subsection illustrate that, similar to a traditional matrix eigenvalue decomposition, the \textit{t-eigenvalue decomposition} in conjunction with the \textit{t-product} enables us to decompose a third order tensor into the product of three third order tensors.  In~\cite{art:Lund18,th:Lund18}, it is shown that traditional functions of matrices can be extended to functions of tensors using the decomposition defined above.  Toward this end, let $\cA \in \mathbb{C}^{n \times n \times \ell}$ and $f(\cdot) : \mathbb{C} \to \mathbb{C}$ be defined on the spectrum of $\circu(\cA)$, and $\cA$ has an eigendecomposition as defined by the \textbf{t-eig}, then the following hold\footnote{A detailed proof can be found in~\cite{art:Lund18} and is omitted here for brevity.}:
\begin{enumerate}
    \item $f(\cA)$ commutes with $\cA$;
    \item $f(\cA^*) = f(\cA)^*$;
    \item $f(\cP * \cA * \cP^{-1}) = \cP * f(\cA) * \cP^{-1}$; and
    \item $f(\cD)*\cP_i = \cP_i*f(\mathbf{d}_i) \; \forall \; i=1, \dots, n$. \\
\end{enumerate}
Using 4), it's easy to show that $f(\cA)$ can be computed as 
\begin{equation}
    f(\cA) = \cP * \left[
    \begin{array}{ccc}
        f(\mathbf{d}_1) & &\\
        & \ddots & \\
        & & f(\mathbf{d}_\ell)\\
    \end{array}
    \right] * \cP^{-1},
\end{equation}
or alternatively, using 3) with eq.~(\ref{eq:D}),\\
$(F_n \otimes I_\ell) f(\circu(\cA)) (F_n^* \otimes I_\ell) = $
\begin{equation}
	 \left[
	 \begin{array}{cccc}
      	f(D_1) &  &  &  \\
      	& f(D_2) & & \\
      	&& \ddots &\\
      	&&& f(D_\ell)
     \end{array}
     \right],
     \label{eq:fD}
\end{equation}
where we note that $f(D_i)$ is defined by the traditional function of a matrix.  Moreover, the product of $f(\cA)$ with some tensor $\cB \in \mathbb{C}^{n \times p \times \ell}$ is computed as
\begin{equation}
    f(\cA)*B = \fld(f(\circu(\cA))*\matvec(\cB)),
\end{equation}
which will become particularly useful when computing solutions to multilinear ordinary differential equations. 

\section{Multilinear System Theory} \label{sec:MLTIsys}
With the preceding definitions of the \textit{t-product}, \textbf{t-eig} and functions of tensors in hand, we are in the position to develop a new approach to multilinear dynamical systems analysis and control.  We proceed by briefly re-stating traditional linear system theory for completeness and presenting a subset of their multilinear extensions.

\subsection{Linear systems theory} \label{subsec:LTI}
In the interest of completeness, we briefly outline a few well known results from traditional linear time-invariant (LTI) system theory.  Consider the system of ordinary differential equations
\begin{eqnarray}
    \dot{\mathbf{x}}(t) & = & A \mathbf{x}(t) + B \mathbf{u}(t) \label{eqn:sys}
\end{eqnarray}
where $\mathbf{x}(t) \in \mathbb{R}^n$, $A \in \mathbb{R}^{n \times n}$ and $B \in \mathbb{R}^{n \times p}$.  Solutions to the system defined in~(\ref{eqn:sys}) are given by
\begin{equation}
    \mathbf{x}(t) = e^{At}\mathbf{x}(0) + \int_0^t e^{A(t-\tau)}B \mathbf{u}(\tau) d \tau,
\end{equation}
where $e^{At}$ is the well known matrix exponential.  Moreover, given the satisfaction of certain Krylov subspace conditions, namely, the controllability matrix $\cC = [B, AB, A^2B, \cdots, A^{(n-1)} B]$ has full row rank (i.e., rank($\cC$) = $n$), the closed-loop eigenvalues of the system in~(\ref{eqn:sys}) can be arbitrarily assigned via control input $\mathbf{u}(t) = -K \mathbf{x}(t)$ with proper choice of $K$, subject to complex eigenvalues appearing as conjugate pairs.  In the sense of stability for LTI systems, we require that the closed-loop eigenvalues $\lambda_i$ of the matrix $(A-BK)$ be contained in the left-half complex plane, i.e., $\{\lambda_i \in \mathbb{C} \; | \; \Re({\lambda_i}) < 0  \}$.

\subsection{From linear to multilinear}
In~\cite{th:Lund18,art:Lund18} it is shown that the zero-input system (a.k.a. the homogeneous system) of multilinear ordinary differential equations (ODEs) given by
\begin{equation}
    \frac{d \cX}{dt}(t) = \dot{\cX}(t) = \cA * \cX(t),
    \label{eq:MLTI_sys}
\end{equation}
has the solution given by
\begin{equation}
    \cX(t) = e^{\cA t} * \cX(0),
    \label{eq:MLTI_sols}
\end{equation}
where $\cX \in \mathbb{R}^{n \times s \times p}$, $\cA \in \mathbb{R}^{n \times n \times p}$, $e^{\cA t}$ is the tensor exponential computed as above with $f(\cA) = e^{\cA t}$ and $*$ is the \textit{t-product}.  It should be noted that within this construct, in conjunction with the definitions outlined in Section~\ref{sec:math_prelims}, the system outlined in~(\ref{eq:MLTI_sys}) can be re-written via the $\fld$ and $\circu$ operators as~\cite{art:Lund18}
\begin{equation}
    \frac{d}{dt} \left[
    \begin{array}{c}
        X^{(1)}\\
        \vdots\\
        X^{(n)}
    \end{array}
    \right] = \circu(\cA) \left[
    \begin{array}{c}
        X^{(1)}\\
        \vdots\\
        X^{(n)}
    \end{array}
    \right],
\end{equation}
with solutions
\begin{equation}
    \left[
    \begin{array}{c}
        X^{(1)}(t)\\
        \vdots\\
        X^{(n)}(t)
    \end{array}
    \right] = e^{\circu(\cA)t} \left[
    \begin{array}{c}
        X^{(1)}(0)\\
        \vdots\\
        X^{(n)}(0)
    \end{array}
    \right],
    \label{eq:bcircexp}
\end{equation}
where we note that the computation of $e^{\circu(\cA)t}$ is performed via the standard matrix exponential.

Using the computation of $e^{\circu(\cA)t}$ obtained in~(\ref{eq:bcircexp}), we can extend the results in~\cite{art:Lund18} to include both the zero-input (homogeneous system) and zero-state solution (forced system).  Indeed, given the multilinear system defined by
\begin{equation}
    \dot{\cX}(t) = \cA * \cX(t) + \cB * \cU(t),
    \label{eq:MLTI_sys_u}
\end{equation}
where $\cX \in \mathbb{R}^{n \times s \times \ell}$, $\cA \in \mathbb{R}^{n \times n \times \ell}$, $\cB \in \mathbb{R}^{n \times q \times \ell}$, and $\cU \in \mathbb{R}^{q \times s \times \ell}$, the solution to such a system is given by
\begin{equation}
    \cX(t) = \underbrace{e^{\cA t} * \cX(0)}_{\text{zero-input}} + \underbrace{\int_0^t e^{\cA (t-\tau)} * \cB * \cU(\tau) d\tau}_{\text{zero-state}}.
    \label{eq:MLTI_U_sols}
\end{equation}

As illustrated in~(\ref{eq:bcircexp}), the zero-input solution can be computed via block-circulant expansion of the tensor $\cA$.  Following the same logic, the zero-state solution can be obtained via similar expansion using the $\fld$ and $\circu$ operators.  Namely,\\
$\displaystyle{\;\; \int_0^t e^{\cA (t-\tau)} * \cB * \cU(\tau) d\tau =}$
\begin{equation}
    \int_0^t \left( e^{\circu(\cA)(t-\tau)}\cdot \matvec(\cB) \right) \cdot \matvec(\cU(\tau)) d\tau,
\end{equation}
where the $\cdot$ notation represents matrix/matrix or matrix/vector multiplication depending on the dimensions of $\cB$ and $\cU$.

\subsection{Stability of the MLTI system}
Evaluating stability of the MLTI system is a bit more complex than evaluating stability of the LTI system.  This stems from the fact that the eigenvalue decomposition of the MLTI system results in a set of eigentuples as opposed to eigenvalues.  Moreover, it's difficult (if not impossible) to define what the negative real-part of the eigentuple $\mathbf{d}_i$ means.  Therefore, we approach stability from construction of the tensor exponential itself as opposed to eigentuple evaluation.  Toward this end, we have the following:\\

\noindent \underline{\textbf{\textit{Claim:}}}  The trajectories $\cX(t)$ of the MLTI system defined in~(\ref{eq:MLTI_sys}) are exponentially stable (i.e., $\cX(t) \rightarrow 0$ as $t \rightarrow 0$ and $|| \cX(t) || \leq e^{-\alpha t}$ for some positive $\alpha$ and $t > 0$) if the eigenvalues of each $D_i$ outlined in~(\ref{eq:D}) have negative real parts.\\~\\

\noindent \underline{\textbf{\textit{Proof:}}}  Using~(\ref{eq:fD}), we can always re-write $e^{\cA t}$ in Fourier space as
\begin{equation}
     e^{\cA t} = (F_n \otimes I_\ell) * e^{\circu(\cA)t} *(F_n^* \otimes I_\ell)
\end{equation}
where each $f(D_i)$ on the right hand side of~(\ref{eq:fD}) is computed as $e^{D_i t}$.  Therefore the trajectories of~(\ref{eq:MLTI_sys}) can also be written as
\begin{equation}
    \cX(t) =(F_n \otimes I_\ell) * e^{\circu(\cA)t} *(F_n^* \otimes I_\ell) * \cX(0).
    \label{eq:alt_MLTI_sols}
\end{equation}
If the eigenvalues of each $D_i$ have negative real part then $f(D_i) = e^{D_i t}$ converges exponentially to the origin as $t \rightarrow \infty$ which implies $e^{\circu(\cA)t}$ and as a result $\cX(t)$ does also. Moreover, because the trajectories are defined by the matrix exponential $e^{D_i t}$, $|| \cX(t) || \leq e^{-\alpha t}$ as $t \rightarrow \infty$ where $\alpha$ is the smallest eigenvalue of all $D_i$, for $i=1,2, \dots,\ell$ \qed\\

For completeness, we provide an alternative stability argument by analyzing the relationship between the eigentuples $\mathbf{d}_i$, and the eigenvalues of the individual $D_i$ matrices outlined in~(\ref{eq:fD}).  Let the eigenvalues associated with $D_i$ be denoted by $\lambda_j^i$ for $i=1,2,\dots,\ell$ and $j=1, 2, \dots, n$.  Then by~(\ref{eq:D}) the first eigentuple $\mathbf{d}_1 = \mathcal{F}^{-1}\{\lambda_1^1, \lambda_1^2, \dots, \lambda_1^\ell\}$ where $\mathcal{F}^{-1}\{\cdot\}$ is the inverse Fourier transform of $\{\cdot\}$, i.e., $\mathbf{d}_1$ is computed as the inverse Fourier transform of the sequence constructed from the first eigenvalue of each $D_i$ (always assumes the eigenvalues are sorted in descending order).  Similarly, the second eigentuple $\mathbf{d}_2= \mathcal{F}^{-1}\{\lambda_2^1, \lambda_2^2, \dots, \lambda_2^\ell\}$ and the $k^\text{th}$ eigentuple is computed as $\mathbf{d}_k= \mathcal{F}^{-1}(\lambda_k^1, \lambda_k^2, \dots, \lambda_k^\ell)$.  In other words, the Fourier transform of the eigentuples $\mathbf{d}_i$, for $i=1,2,\dots, \ell$ are the collection of eigenvalues of the $D_i$, for $i=1,2,\dots, \ell$, where $\mathcal{F}\{ \mathbf{d}_1 \}$ produces the first (largest) eigenvalue $\lambda_1^i$ in each $D_i$, $\mathcal{F}\{ \mathbf{d}_2 \}$ produces the second eigenvalue $\lambda_2^i$, etc. $\implies$ the complex-plane maps to the eigentuples $\mathbf{d}_i$ through the inverse Fourier transform.  Therefore, for stability,  we require $\{ \mathcal{F} \{  \mathbf{d}_i \} \in \mathbb{C} \; | \; \Re{\{\mathcal{F} \{  \mathbf{d}_i \}\}} < 0 \}$ for $i = 1,2, \dots, \ell$.\\

\subsection{Feedback control and eigentuple re-assignment}
With the notion of stability for MLTI systems in hand, we now present a method to control the system through multilinear state feedback.  Similar to traditional linear state feedback, prior to designing a state feedback control law we require certain controllability conditions.  Namely, we need to impose a rank condition on the controllability tensor $\hat{\cC}$ constructed as
$$
    \hat{\cC} = \left[\cB, \cA*\cB, \cA^2*\cB, \cdots, \cA^{n-1} * \cB \right],
$$
where $\cA \in \mathbb{R}^{n \times n \times \ell}$, $\cB \in \mathbb{R}^{n \times q \times \ell}$ and $*$ is the \textit{t-product}.  The rank condition we seek was defined in~\cite{KilMP08,KilM09,Bra10} (referred to as ``\textit{tubal-rank}") and stems from the fact that evaluating a zero eigentuple (or singular tuple) is fundamentally different than evaluating an eigenvalue (or singular value).  Toward this end, the tubal rank of a tensor is defined as:\\

\noindent \textbf{\textit{tubal-rank~\cite{KilMP08,KilM09,Bra10}:}} Suppose $\mathbf{a} \in \mathbb{R}^{1 \times 1 \times n}$ is a tubal scalar. Then its tubal-rank is the number of its non-zero Fourier coefficients.  If its tubal-rank is $n$, we say it is \textbf{\textit{invertible}}, if it is less than $n$, it is not. In particular, the tubal-rank is 0 iff $\mathbf{a} = 0$.

To ensure controllability of the MLTI system, we need all singular tuples of $\hat{\cC}$ to be non-zero, where the singular tuples of a tensor are computed similar to the eigentuples~\cite{KilMP08,KilM09,Bra10}.  Alternatively, we can check the controllability by evaluating the rank of the block controllability matrix
\begin{eqnarray}
    \circu(\hat{\cC}) & = & \left[\cB_v, \cA_c \cdot \cB_v, \cA_c^2 \cdot \cB_v, \cdots, \cA_c^{n-1} \cdot \cB_v \right],
    \label{eq:tctrb}
\end{eqnarray}
where we define $\cB_v = \matvec(\cB)$ and $\cA_c = \circu(\cA)$ for notational convenience.  For complete controllability, we require $\text{rank}(\circu(\hat{\cC})) = \ell n$.

\noindent \textbf{\textit{MLTI State Feedback (theory):}} Given the MLTI system defined in~(\ref{eq:MLTI_sys}), an assuming the controllability condition defined above is satisfied, then using the control input $\cU(t) = -\cK * \cX(t)$, where $\cK \in \mathbb{R}^{q \times n \times \ell}$, we can place the closed-loop eigentuples $\hat{\mathbf{d}}_i$ arbitrarily as long as the complex elements of $\hat{\mathbf{d}}_i$ are assigned in conjugate pairs.  As a result, the trajectories of the new system will satisfy the multilinear ODEs given by
\begin{equation}
    \dot{\cX}(t) = \left( \cA - \cB*\cK \right) * \cX(t),
    \label{eq:MLTI_sys_u_control}
\end{equation}
where the eigentuples of the feedback tensor $\left( \cA - \cB*\cK \right)$ are arbitrarily assigned.  

\noindent \textbf{\textit{MLTI State Feedback (construction):}} While the above theory is technically sound, in practice choosing the feedback tensor $\cK$ such that the closed-loop tensor $\left( \cA - \cB*\cK \right)$ has a desired set of eigentuples is challenging.  This partially stems from the fact that characteristic polynomials for tensors and/or companion forms are still ongoing research efforts by the authors.  Therefore, using traditional approaches borrowed from linear systems theory (casting to control canonical form through a similarity transformation or developing a desired characteristic polynomial) fall short.  As a result, we turn once again to~(\ref{eq:D}) and note that the $D_i$ on the right hand side of~(\ref{eq:D}) contain exactly the spectrum of $\circu(\cA)$ and by definition (although trough a Fourier transform mapping) determine the eigentuples $\mathbf{d}_i$ of $\cA$.  As a result, rather than attempting to re-assign the eigentuples directly, in practice, it's more convenient to assign the eigenvalues of each $D_i$ through traditional linear systems theory, i.e., construct the new matrix $\left( D_i - B_i \cdot K_i \right)$ for $i=1,2,\dots,\ell$, and map those eigenvalues to the desired closed-loop eigentuples $\hat{\mathbf{d}}_i$ through the Fourier transform\footnote{Here we note that $B_i \in \mathbb{R}^{n \times q}$ is the first block of $\matvec({\cB})$}.

\section{Illustrative Example} \label{sec:example}
To help solidify the theory developed in Section~\ref{sec:MLTIsys}, we present an illustrative example here.  Consider the system of~(\ref{eq:MLTI_sys_u}) given as
\begin{equation}
    \dot{\cX}(t) = \cA * \cX(t) + \cB * \cU(t),
    \label{eq:example_eq}
\end{equation}
with $\cA \in \mathbb{R}^{2 \times 2 \times 2}$ who's frontal slices are given by
\begin{equation}
    \cA^{(1)} = \left[
    \begin{array}{cc}
         -6 & 6  \\
         -10 & 0 
    \end{array}
    \right] \text{ and } \cA^{(2)} = \left[
    \begin{array}{cc}
         0 & 2  \\
         8 & 2 
    \end{array}
    \right],
\end{equation}
and $\cB \in \mathbb{R}^{2 \times 1 \times 2}$ who's frontal slices are given by
\begin{equation}
    \cB^{(1)} = \cB^{(2)} = \left[
    \begin{array}{c}
         1  \\
         1 
    \end{array}
    \right],
\end{equation}
and our \textit{state-matrix} $\cX(t) \in \mathbb{R}^{2 \times 1 \times 2}$ (analogous to a state-vector) is given as
\begin{equation}
    \cX^{(1)}(t) \left[
    \begin{array}{c}
         x_1(t)  \\
         x_2(t) 
    \end{array}
    \right] \text{ and } \cX^{(2)}(t) \left[
    \begin{array}{c}
         x_3(t)  \\
         x_4(t) 
    \end{array}
    \right].
\end{equation}

\begin{figure}[h!]
\begin{center}
\includegraphics[width=0.9\columnwidth]{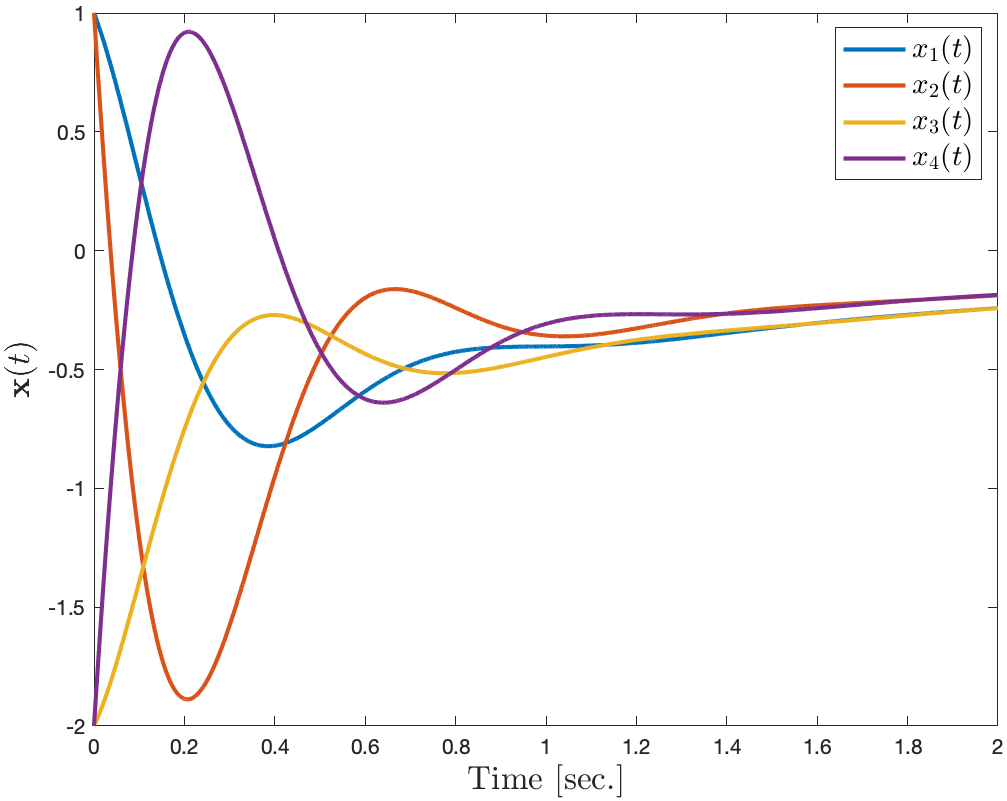}
\end{center}
\vspace{-2ex}
\caption{\label{fig:open_loop}Illustration of the open-loop trajectories of the system defined in~(\ref{eq:example_eq} with no control applied, i.e., $\cU(t) = 0$.).  
}
\end{figure}

\begin{figure}[h!]
\begin{center}
\includegraphics[width=0.9\columnwidth]{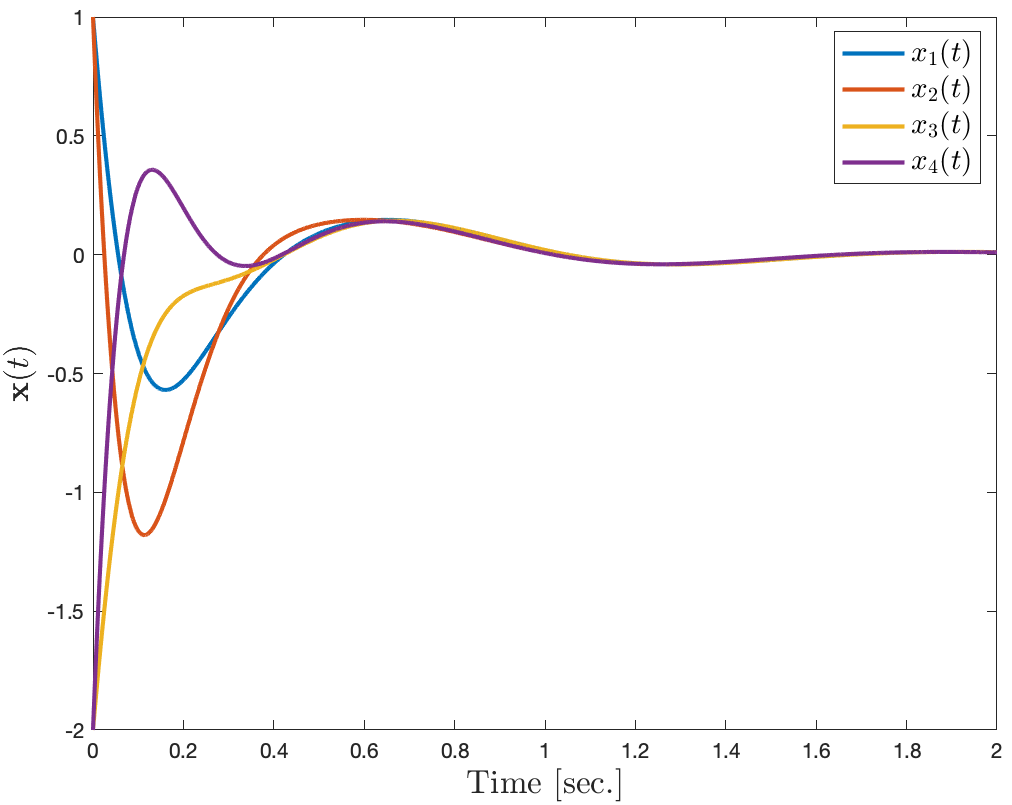}
\end{center}
\vspace{-2ex}
\caption{\label{fig:closed_loop}Illustration of the closed-loop trajectories of the system defined in~(\ref{eq:example_eq} with the control input $\cU(t) = -\cK * \cX(t)$.  
}
\end{figure}

Performing the Fourier transform on the eigentuples returned by $\textbf{t-eig}(\cA)$ yields open-loop eigentuples (governing the system trajectories - i.e., the eigenvalues of the $D_i$ matrices outlined in~(\ref{eq:D})) at
$$
    \bar{\mathbf{d}}_1 = \left[
    \begin{array}{c}
         -3.414  \\
         -0.586
    \end{array}
    \right] \text{ and } 
    \bar{\mathbf{d}}_2 = \left[
    \begin{array}{c}
         -4 + j7.07  \\
         -4 - j7.07
    \end{array}
    \right],
$$
where we've used $\bar{\mathbf{d}}_i$ to denote that $\mathbf{d}_i \in \mathbb{R}^{1 \times 1 \times 2}$ is a tubal-scalar.  Although the open-loop system is stable, we aim to perform eigenvalue re-assignment to improve the closed-loop characteristics.  Toward this end, using~(\ref{eq:D}) we compute the matrices $D_i$ to be
\begin{equation*}
    D_1 = \left[
    \begin{array}{cc}
         -6 & 7  \\
         -2 & 2 
    \end{array}
    \right] \text{ and } D_2 = \left[
    \begin{array}{cc}
         -6 & 3  \\
         -18 & -2 
    \end{array}
    \right],
\end{equation*}
and desire our closed loop eigenvalues of each $D_i$ to be placed at $\lambda_{1,2}^1 = -2 \pm j5$ and $\lambda_{1,2}^2 = -10 \pm j10$, resulting in $K_1 = [27 \;\; -27]$ and $K_2 = [16.35 \;\; -4.35]$.  Finally, ``stacking" each $K_i$ into a tensor $\bar{\cK}$ as $\bar{\cK}^{(1)} = K_1$ and $\bar{\cK}^{(2)} = K_2$ and performing the Fourier transform on $\bar{\cK}$ yields 
\begin{equation*}
    \cK^{(1)} = \left[
    \begin{array}{cc}
         43.35 &  -31.35  \\
    \end{array}
    \right] \text{ and } \cK^{(2)} = \left[
    \begin{array}{cc}
         10.64 &  -22.64\\
    \end{array}
    \right],
\end{equation*}
and letting $\cU(t) = -\cK * \cX(t)$ produces the desired closed-loop response.  The trajectories for both the open-loop system and closed-loop system are illustrated in Figures~\ref{fig:open_loop} and~\ref{fig:closed_loop} respectively.  
\section{Conclusions and Future Directions} \label{sec:conclusions}
This paper presented a new approach to analysis and design of multilinear systems theory and control.  The approach is based on a recently developed tensor product and tensor eigenvalue decomposition that lays the foundation for solutions to multilinear dynamical systems through the definition of a tensor-exponential.  Using this formulation, we extend traditional linear systems theory and control to their multilinear counterparts.  Namely, we introduce new notions of stability, controllability, and state-feedback of multilinear dynamical systems.  

We note that using the above definitions for decomposing third order tensors is still very immature.  While there are many different research directions to investigate within this framework, our immediate focus will be on developing a notion of a characteristic polynomial for tensors which leads to defining what it means for a tensor to be in companion form.  Moreover, we wish to investigate the observability conditions generally present in linear systems and define a framework for multilinear observability as well as multilinear state estimation.  Finally, we wish to explore application areas for such a framework that may arise the real-world problems.



\bibliographystyle{IEEEtran}
\bibliography{ACC}

\end{document}